\documentclass[10pt, a4paper]{article}

\usepackage[final]{lrec-coling2024} % this is the new style
\usepackage{multirow}
\usepackage{graphicx}
\usepackage{booktabs}
\usepackage{amsmath}
\usepackage{authblk}
\usepackage[stable]{footmisc}

\title{
    Denoising Table-Text Retrieval\\
    for Open-Domain Question Answering
}

\name{Deokhyung Kang$^{1}$, Baikjin Jung$^{2}$, Yunsu Kim$^{3}$, Gary Geunbae Lee$^{1,2}$} 

\address{$^{1}$Graduate School of Artificial Intelligence, POSTECH, Republic of Korea, \\
        $^{2}$Department of Computer Science and Engineering, POSTECH, Republic of Korea \\
        $^{3}$aiXplain, Inc. Los Gatos, CA, USA, \\
         \{deokhk, bjjung, gblee\}@postech.ac.kr, yunsu.kim@aixplain.com\\}

\abstract{
In table-text open-domain question answering, a retriever system retrieves relevant evidence from tables and text to answer questions. Previous studies in table-text open-domain question answering have two common challenges: firstly, their retrievers can be affected by false-positive labels in training datasets; secondly, they may struggle to provide appropriate evidence for questions that require reasoning across the table.
To address these issues, we propose \textbf{D}en\textbf{o}ised \textbf{T}able-\textbf{Te}xt \textbf{R}etriever (DoTTeR). Our approach involves utilizing a denoised training dataset with fewer false positive labels by discarding instances with lower question-relevance scores measured through a false positive detection model. Subsequently, we integrate table-level ranking information into the retriever to assist in finding evidence for questions that demand reasoning across the table. To encode this ranking information, we fine-tune a rank-aware column encoder to identify minimum and maximum values within a column. Experimental results demonstrate that DoTTeR significantly outperforms strong baselines on both retrieval recall and downstream QA tasks. Our code is available at \url{https://github.com/deokhk/DoTTeR}. \\ \newline  }

\begin{document}

\maketitleabstract

\section{Introduction}

In open-domain question answering (ODQA), a \textit{retriever} is a system that brings evidence supporting potential answers to the given question from an information source.
These pieces of evidence are then used by a \textit{reader} system, which answers the question in effect. Typically, one would expect the evidence to consist solely of text, but in practice, it is highly likely that it also contains images or \textbf{tables}, necessitating ODQA models to perform multi-hop aggregation of different modalities of information.
For example, the OTT-QA~\citeplanguageresource{chen2021ottqa} dataset sets up a situation where ODQA models must use both tables and text to answer the given question.

This setting presents practical challenges for conventional retrievers~\cite{karpukhin2020dense,qu-etal-2021-rocketqa}, as assessing the relevance to the question solely based on a single modality often results in incomplete measurement. At the same time, the size of the table frequently surpasses the token limit of standard pretrained language models~\cite{devlin-etal-2019-bert,liu2019roberta}. To address these challenges, \textbf{fusion retrieval} is presented~\citeplanguageresource{chen2021ottqa}. They first pre-align a row in a table to their related passages using entity linking, forming a "\textbf{fused block}". Then, the retriever identifies relevant fused blocks, and the reader model extracts the answer from the concatenated fused blocks.

\begin{figure}[!t]
    \centerline{\includegraphics[width=\columnwidth]{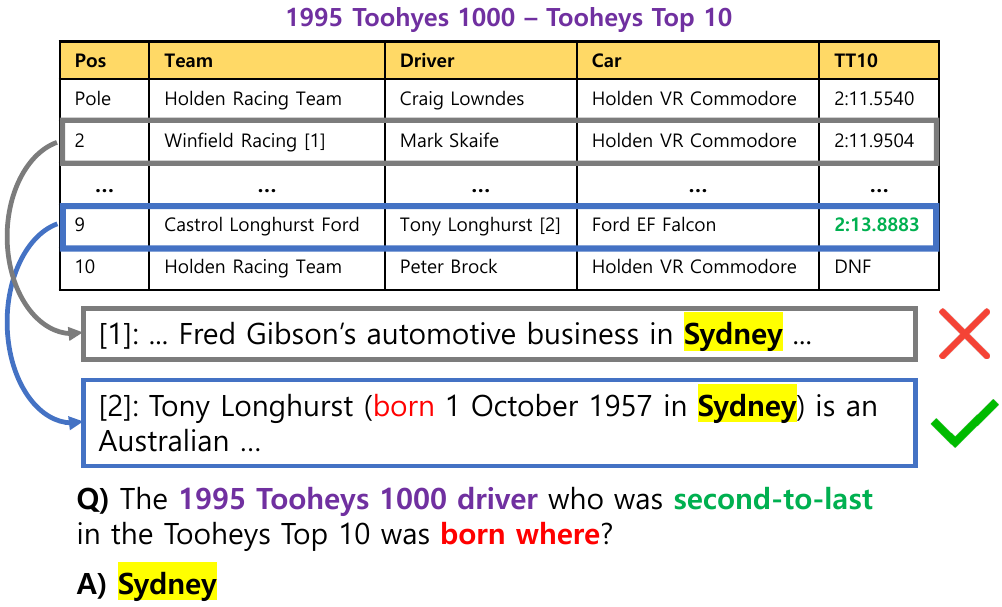}}
    \caption{An example of a question and related table in OTT-QA. Two fused blocks contain the answer "Sydney" to the question, but only the blue-bordered block satisfies the conditions required by the question. }
    \label{figure_1}
\end{figure}

While fusion retrieval successfully addresses the aforementioned challenges, it still has two limitations.

\textbf{1) False positive instances for retriever.} When training the retriever, the fusion retrieval treats all fused blocks containing answers relevant to the question, as block-level supervision is unavailable in OTT-QA. This leads to false-positive instances, introducing \textbf{noise} during training. Figure~\ref{figure_1} depicts two fused blocks. One is bordered in gray, and the other is bordered in blue. Both blocks contain the answer (Sydney) to the question, but only the block bordered in blue is relevant, as this block contains the second-to-last row in the Tooheys Top 10.

\textbf{2) Neglect to utilize table-level information.} Certain questions in OTT-QA require information beyond the scope of a fused block to answer. For instance, to answer the question in Figure~\ref{figure_1}, the fusion retriever should identify the fused block containing the second-to-last row in the Tooheys Top 10. For this purpose, the retriever should process table-level ranking information across fused blocks. However, the fusion retriever lacks access to such information since the fused block is confined to row-level information. Lack of table-level information leads to the retriever training with incomplete features, causing \textbf{additional noise} in the training process.

In this paper, we propose the \textbf{D}en\textbf{o}ised \textbf{T}able-\textbf{Te}xt \textbf{R}etriever (DoTTeR), built upon the state-of-the-art fusion retrieval model OTTeR~\cite{huang-etal-2022}, to address the above problems. Our approach comprises two main components: \textbf{(1) Denoising OTT-QA.} We train a false positive detection model that measures question-fused block relevance scores to de-noise the training dataset. We use this model to eliminate potential false positive instances for the retriever by only keeping the block with the highest relevance score for each question. \textbf{(2) Rank-Aware Table Encoding (RATE).} RATE involves a rank-aware encoder that is fine-tuned for locating the minimum and maximum values in numeric columns of a given table. We use the encoder to provide OTTeR with a dense representation of a given table and expect the retriever to replenish such information beyond the scope of a block as a result. Experimental results on OTT-QA show that DoTTeR significantly improves performance in both table-text retrieval and downstream question-answering tasks.

\section{Methods}

Although a typical ODQA retriever aims to identify fused blocks related to the given question, table-text QA datasets~\citeplanguageresource{chen-etal-2020-hybridqa, chen2021ottqa} including OTT-QA do not pair questions with their corresponding fused blocks as annotating the answer row in a table is costly.
Previous studies~\cite{chen2021ottqa, zhong-etal-2022, huang-etal-2022} address this issue by considering all fused blocks containing the answer entity as relevant to the question. However, this supposition is susceptible to noisy labeling in that there can be \textbf{false-positive blocks} if the answer is a `common' entity appearing in several table rows. 

In addition, those previous studies use two encoders for dense retrieval, where a question ${q}$ and a fused block ${b}$ are separately encoded into ${d}$-dimensional vectors by the question encoder $E_Q$ and the block encoder $E_B$, respectively.
Here, the similarity between the question and the fused block 
\begin{equation*}
s(q, b) = E_Q(q)^T \cdot E_B(b)
\end{equation*}
is their dot product.

However, $E_B(b)$ contains only the information of a single fused block $b$, not capturing `table-level' information beyond the block's scope.
Specifically, our interest is the \textbf{rank} of numerical values belonging to the same column because table--text reasoning often involves the superlative operation over a column, e.g., "the earliest Olympic event"~\citeplanguageresource{chen-etal-2020-hybridqa}.

\begin{figure*}[!ht]
    \centerline{\includegraphics[width=\textwidth]{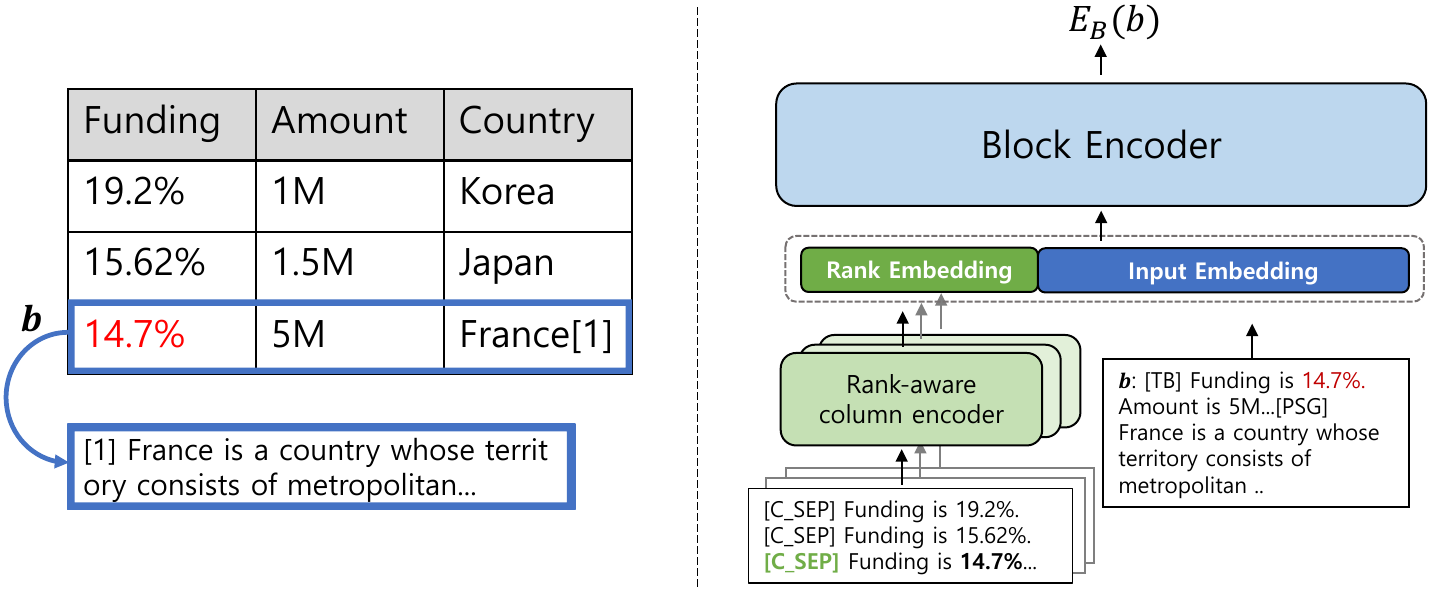}}
    \caption{An overview of the encoding process for a fused block $b$ with RATE. The fused block $b$ belongs to the table on the left and is encoded into $E_{B}(b)$ from the concatenation of the rank embedding, extracted from the rank-aware column encoder, and the input embedding.}
    \label{figure_2}
\end{figure*}

\subsection{Denoising OTT-QA}

Inspired by \cite{lei-etal-2023-s3hqa, kumar-etal-2023-multi}, we address the problem of false-positive training instances for the retriever by employing a false-positive detection model. This model is designed to identify false-positive blocks by taking a question concatenated with a single fused block as input and outputs a question-block relevance score $s$. We implement this model by training a BERT~\cite{devlin-etal-2019-bert} model with a single linear layer on a binary classification task. As we need noiseless relevant and irrelevant fused blocks for given questions to train this model, we divide the OTT-QA dataset, denoted as $D$, into two partitions. $D$ comprises instances represented as $<q, a, t, B>$,  where $q$ is a question, $a$ is an answer, $t$ is a corresponding table, and $B$ is a set of corresponding fused blocks to t. We partition $D$ into two categories: $D_1$ and $D_{2+}$. $D_1$ comprises instances without noise, featuring only one fused block $(b+)$ from $B$ contains $a$ (exact match on text) , while $D_{2+}$ includes instances with multiple fused blocks from B contain $a$. We use $D_1$\footnote{We found that approximately 63.3\% of the training instances belong to $D_{1}$, while the remaining 37.7\% of the training instances belong to $D_{2+}$.}  to train the false-positive detection model.

During training, the model treats <$q, b^{+}$> as a positive instance and $<q, b^{-}>$ as a negative instance, where $b^{-}$ is a fused block with the highest BM25 ~\cite{robertson2009probabilistic} score to $q$ among the blocks in the subset of $B$ without containing $a$. The training process involves minimizing the binary cross-entropy (BCE) loss as follows:
$$Loss^{bce} = BCE(s, y)$$
where $y \in \{0, 1\}$ is the label of the fused block indicating whether the block is relevant to the given question or not.
Then, we remove potentially false-positive fused blocks from $D_{2+}$ by only retaining a fused block with the highest question-block relevance score for each question.

\subsection{Rank-Aware Table Encoding (RATE)}

To provide ranking information to the block encoder, we propose \textbf{RATE} (\textbf{R}ank-\textbf{A}ware \textbf{T}able \textbf{E}ncoding), which leverages a rank-aware column encoder $R$ to incorporate ranking information while encoding fused blocks. 
\paragraph{Training the rank-aware column encoder.} Consider a list of values [19.2\%, 15.62\%, 14.7\%] under the 'Funding' column header in Figure~\ref{figure_2}. These values are linearized into text format as '[C\_SEP] Funding is 19.2\% [C\_SEP] Funding is 15.62\% [C\_SEP] Funding is 14.7\%', where [C\_SEP] is a special token representing the rank information of the nearest column value.

We train the rank-aware column encoder $R$ to embed ranking information into these token representations. This involves adding two linear layers—max and min—on top of $R$ during training. These layers assign probabilities to each token, indicating whether it represents the maximum or minimum value among the inputs, respectively. To train these layers with $R$, we set a label for each token to 1 if the token's index corresponds to the [C\_SEP] token associated with the max/min value; otherwise, it is set to 0. The training process minimizes the cross-entropy loss between the predicted probabilities and the label. Once the training is done, these layers are removed, allowing the RATE module to output rank embeddings for each token.

\paragraph{Incorporating the ranking information.}To incorporate ranking information for table values within a fused block during encoding, we divide the original table, to which the fused block belongs, into columns. The rank-aware column encoder processes a list of values from each column as input and generates a list of rank embeddings.
We take rank embeddings corresponding to table values within the fused block and feed it along with an input embedding to the block encoder during encoding. Figure~\ref{figure_2} illustrates such a process. 

\paragraph{Training the retriever.} We train the question encoder and the block encoder to maximize the similarity between the question and the relevant block while keeping the rank-aware column encoder frozen. Following ~\citet{huang-etal-2022}, we assign a positive block $b^+$ and $m$ negative blocks $\{{b^-_i}
\}_{i=1}^m$ for given question $q$ and minimize the cross-entropy loss $L$:
$$L(q, b^+, b^-_{1}, ... ,b^-_{m})= -
\log\frac{e^{s(q,b^+)}}{{e^{s(q,b^+)}}+
\sum_{i=1}^{m}e^{s(q,b^-_{i})}}
$$
\begin{table*}[!ht]
\resizebox{\textwidth}{!}{%
\begin{tabular}{l|cccc|cccc}
\hline
\multirow{2}{*}{\textbf{Methods}} & \multicolumn{4}{c|}{\textbf{Block Recall}}                    & \multicolumn{4}{c}{\textbf{Table Recall}}                     \\ \cline{2-9} 
                                  & R@1           & R@10          & R@15          & R@20          & R@1           & R@10          & R@15          & R@20          \\ \hline
BM25                              & 23.7          & 45.3          & 47.9          & 50.0          & 32.8          & 62.1          & 65.4          & 67.9          \\ \hline
Bi-Encoder~\cite{kostic-etal-2021}                        & -             & -             & -             & -             & 46.2          & 70.9          & -             & 76.0          \\
Tri-Encoder~\cite{kostic-etal-2021}                      & -             & -             & -             & -             & 47.7          & 70.8          & -             & 77.7          \\
CARP~\cite{zhong-etal-2022}                              & 16.3          & 46.7          & -             & -             & 49.0          & 74.0          & -             & -             \\
OTTeR*~\cite{huang-etal-2022}                            & 31.1          & 66.7          & 72.6          & 75.6          & \textbf{57.6} & 80.9          & 83.8          & 85.2          \\ \hline
\textbf{DoTTeR (ours)}            & \textbf{37.6} & \textbf{70.4} & \textbf{74.1} & \textbf{76.6} & 57.3          & \textbf{83.6} & \textbf{85.8} & \textbf{87.5} \\
\textit{\quad w/o denosing OTT-QA}      & 31.7          & 68.0          & 72.6          & 74.0   & 57.3          & 82.7          & 84.8          & 85.9          \\
\textit{\quad w/o RATE}                 & 34.9          & 68.1          & 72.4          & 75.3          & 55.0          & 82.1          & 83.8          & 86.7          \\ \hline
\end{tabular}%
}
\caption{Retrieval results on  OTT-QA dev set. Note that we directly copy the reported results from the papers and leave the blanks if they were not reported. * denotes results reproduced by us.}
\label{tab:retrieval}
\end{table*}
\section{Experiments}
\subsection{Dataset and Evaluation Metrics}
We evaluate our system on table-text retrieval and downstream question-answering tasks using OTT-QA~\citeplanguageresource{chen2021ottqa}, a large-scale English ODQA dataset over tables and text. The dataset is the sole benchmark within the domain of ODQA over tables and text and has 42K / 2K / 2K questions for the train/dev/test set, respectively. Additionally, it offers a corpus consisting of over 410K tables and 6.3M passages from Wikipedia. We utilize preprocessed fused blocks from OTTeR~\cite{huang-etal-2022}, where BLINK~\cite{wu-etal-2020-scalable} was employed as the entity linker. This results in 5.4M fused blocks.

We evaluate retrieval performance using table recall@k and block recall@k metrics. Table recall@k measures the percentage of questions in the evaluation set for which at least one of the top-k retrieved fused blocks belongs to the ground truth table. Block recall@k is a coarse-grained metric that measures the percentage of questions in the evaluation set for which at least one of the top-k retrieved fused blocks belongs to the ground truth table and contains the answer.

For question answering, we employ EM (Exact Match) and F1 score metrics to evaluate performance.

\subsection{Implementation Details\footnote{Unless otherwise specified, we utilize the OTT-QA training split for model training.}}
\paragraph{False-Positive Detection.} We initialize the model's encoder with BERT-base-cased ~\cite{devlin-etal-2019-bert} and train it for 5 epochs with a batch size of 32 and a learning rate of 2e-5. The training process took 1 hour, utilizing two NVIDIA GeForce RTX 3090 GPUs.
\paragraph{Rank-Aware Table--Text Retriever.} We first train the Rank-Aware Column Encoder. This involves extracting 626,774 numerical columns with values such as numbers or dates from the table corpus provided by OTT-QA. After initializing the model with RoBERTa-base~\cite{liu2019roberta}, we then train the model for 60,000 steps with a batch size of 32 and a learning rate of 5e-5. The training process took 6 hours, utilizing four A100-80GB GPUs.  

Then, we initialize both the question and block encoder with the synthetic pretrained checkpoint released from OTTeR~\cite{huang-etal-2022}. We train both encoders for 20 epochs with a batch size of 64 and a learning rate of 2e-5. The training process took 26 hours on four A100-40GB GPUs.

\paragraph{Cross-Block Reader (CBR).}
Following ~\citet{huang-etal-2022}, We adopt the cross-block reader (CBR) as our reader model. This model extracts the best answer span from the concatenated top 15 retrieved fused blocks. We use Longformer-base~\cite{Beltagy2020Longformer} as the backbone of the reader. We train the reader for 5 epochs with a batch size of 16 and a learning rate of 1e-5. The training process took 30 hours on four A100-40GB GPUs.

\section{Results and Analysis}

We evaluate DoTTeR by comparing it with various retrieval methods, including the sparse retrieval method BM25~\cite{robertson2009probabilistic}, dense retrieval method~\cite{kostic-etal-2021}, and fusion retrieval method~\cite{zhong-etal-2022, huang-etal-2022} including OTTeR~\cite{huang-etal-2022}, the state-of-the-art fusion retrieval model for table-text retrieval.
\paragraph{Main results.} Table~\ref{tab:retrieval} presents the retrieval results comparing DoTTeR with other baselines. The results highlight that our method significantly outperforms the baselines in block and table recall on the OTT-QA development set, particularly when k is small. Compared to OTTeR, DoTTeR notably enhances block recall, achieving a substantial 6.5\% gain in block recall@1. This demonstrates the efficacy of the proposed false-positive removal and RATE in improving retrieval at a fine-grained level, especially relevant for QA. This highlights the effectiveness of DoTTeR as a retrieval model for table-text Open-Domain Question Answering (ODQA).

\paragraph{Ablation studies.} We conduct ablation studies to investigate the effect of proposed methods on table-text retrieval. Firstly, we investigate the effect of denoising. For \textit{w/o denoising OTT-QA}, we use the original OTT-QA data for training. This leads to a significant drop in retrieval recall, even falling below OTTeR in block recall@20. This drop underscores the importance of denoising OTT-QA and emphasizes its essential role as a preliminary step before integrating the ranking information. Subsequently, we assess the effect of RATE. For \textit{w/o RATE}, we do not provide rank information to the block encoder and use denoised OTT-QA data for training. This also leads to a substantial drop in retrieval recall, highlighting the effectiveness of RATE.

\begin{figure}[!h]
    \centerline{\includegraphics[width=\columnwidth]{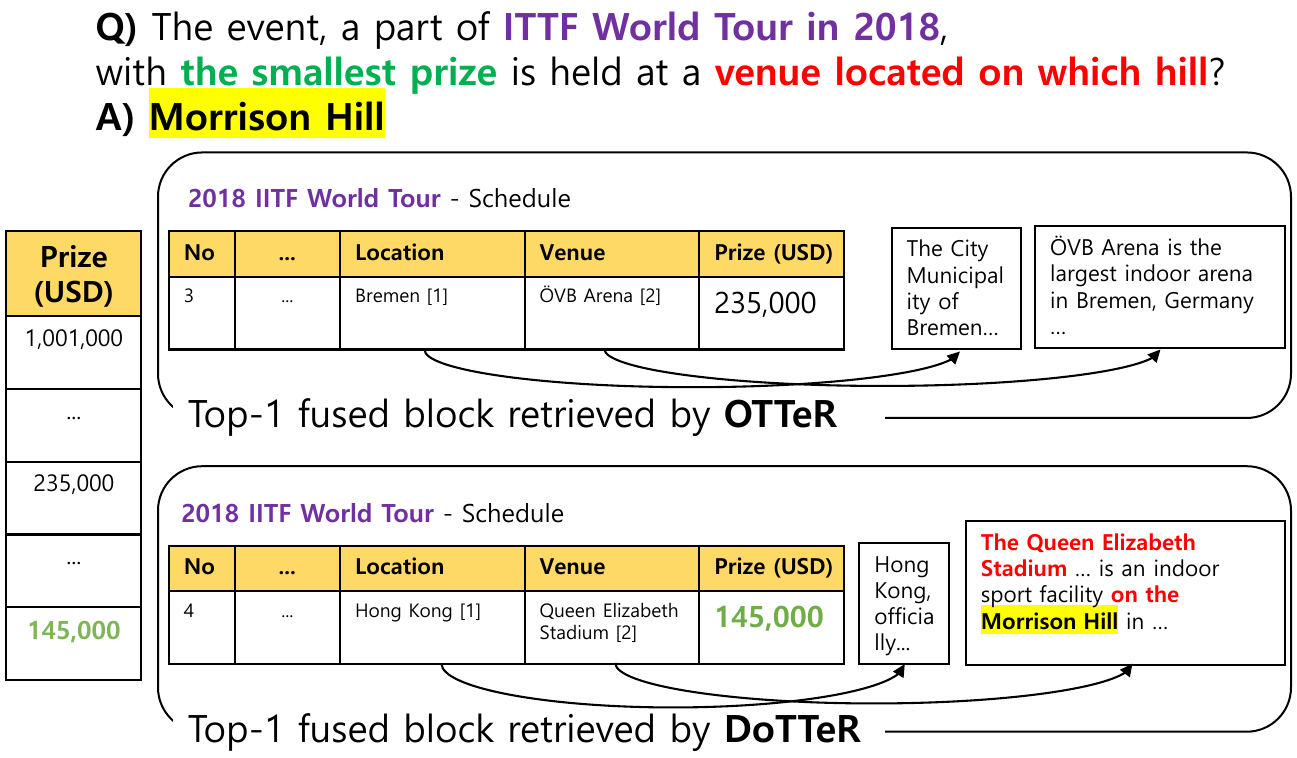}}
    \caption{
        Top-1 fused blocks retrieved by OTTeR and DoTTeR, respectively.
    }
    \label{figure_3}
\end{figure}
\paragraph{Case study.} To demonstrate the DoTTeR's effective utilization of ranking information, we provide an example of a top-1 fused block retrieved by both DoTTeR and OTTeR. To answer the question in Figure~\ref{figure_3}, the retriever model should retrieve the fused block with the lowest rank for the prize field from the relevant table.
As depicted in Figure~\ref{figure_3}, DoTTeR accomplishes this task, retrieving the relevant fused block with the lowest rank. However, OTTeR retrieves the fused block associated with the relevant table but fails to obtain the fused block with the lowest rank. This result demonstrates that RATE facilitates the retrieval of evidence for questions requiring ranking information.

\begin{table}[!h]
\resizebox{\columnwidth}{!}
{%
\begin{tabular}{l|cc|cc}
\hline
\multirow{2}{*}{\textbf{Methods}} & \multicolumn{2}{c|}{\textbf{Dev}} & \multicolumn{2}{c}{\textbf{Test}} \\
\cline{2-5} & \multicolumn{1}{c|}{EM} & F1 & \multicolumn{1}{c|}{EM} & F1 \\ \hline
BM25 + HYBRIDER~\citeplanguageresource{chen-etal-2020-hybridqa}                                      & \multicolumn{1}{c|}{10.3}          & 13.0                             & \multicolumn{1}{c|}{9.7}           & 12.8                            \\
BM25 + DUREPA~\cite{li-etal-2021-dual}                                       & \multicolumn{1}{c|}{15.8}          & -                                & \multicolumn{1}{c|}{-}             & -                               \\
Iterative-Retrieval + CBR ~\citeplanguageresource{chen2021ottqa}                           & \multicolumn{1}{c|}{14.4}          & 18.5                             & \multicolumn{1}{c|}{16.9}          & 20.9                            \\
Fusion-Retrieval + CBR ~\citeplanguageresource{chen2021ottqa}                              & \multicolumn{1}{c|}{28.1}          & 32.5                             & \multicolumn{1}{c|}{27.2}          & 31.5                            \\
OTTeR + CBR* ~\cite{huang-etal-2022}                                         & \multicolumn{1}{c|}{35.8}          & 41.5                             & \multicolumn{1}{c|}{34.8}          & 40.7                            \\ \hline
\textbf{DoTTeR + CBR (ours)}                          & \multicolumn{1}{c|}{\textbf{37.8}} & \textbf{43.9}                    & \multicolumn{1}{c|}{\textbf{35.9}} & \textbf{42.0}                   \\
\textit{\quad w/o denoising OTT-QA + CBR}                  & \multicolumn{1}{l|}{37.1}   & \multicolumn{1}{l|}{43.0} & \multicolumn{1}{l|}{35.5}   & \multicolumn{1}{l}{41.5} \\
\textit{\quad w/o RATE + CBR}                              & \multicolumn{1}{c|}{35.8}          & 41.8                             & \multicolumn{1}{c|}{35.1}          & 41.0                            \\ \hline
\end{tabular}%
}
\caption{QA results on OTT-QA dev and blind test set. * denotes results reproduced by us.}
\label{tab:qa}
\end{table}

\paragraph{Question Answering results.} We implement an open-domain QA system using DoTTeR as a retriever and CBR as a reader. We compare the system with other baselines consisting of retriever and reader.
Table~\ref{tab:qa} shows that our method outperforms existing openQA systems, with notable EM and F1 gains. These results demonstrate that improved table-text retrieval of the proposed method leads to improvements in downstream QA. 

\section{Related Work}
Previous studies on table-text ODQA can be classified into two primary approaches. ~\citetlanguageresource{chen2021ottqa} proposed the early-fusion approach (fusion retrieval), wherein they pre-align the table segments to their related passages, forming fused blocks. Subsequently, they retrieved the top K fused blocks and fed them into the reader model, a long-range transformer model \cite{ainslie-etal-2020-etc}. Several studies~\cite{zhong-etal-2022, huang-etal-2022, park2023rink} follow this approach.
On the other hand, \cite{ma-etal-2022,ma-etal-2023} employed a late-fusion approach. They linked the table segments and relevant passages after retrieval. Our work follows the early-fusion strategy, which is lightweight and practical. We extend this approach by tackling the limitations of the fusion retrieval models.

\section{Conclusion}
In this paper, we proposed \textbf{D}en\textbf{o}ised \textbf{T}able-\textbf{Te}xt \textbf{R}etriever (DoTTeR) to address issues of 
false-positive training instances for retrievers and neglecting table-level information in previous table-text retrieval systems. To mitigate these issues, we train the false-positive detection model with noiseless instances from OTT-QA and utilize this model to denoise the dataset. Additionally, we incorporate table-level ranking information into the retriever through rank-aware table encoding (RATE). Experimental results demonstrate that our approach significantly improves retrieval recall and downstream question-answering performance. 

\section{Acknowledgments}
This research was supported by the MSIT(Ministry of Science and ICT), Korea, under the ITRC(Information Technology Research Center) support program(IITP-2024-2020-0-01789) supervised by the IITP(Institute for Information \& Communications Technology Planning \& Evaluation) and by Institute of Information \& communications Technology Planning \& Evaluation (IITP) grant funded by the Korea government(MSIT) (No.2019-0-01906, Artificial Intelligence Graduate School Program(POSTECH)). We thank the anonymous reviewers for their helpful
comments and suggestions.
\nocite{*}
\section{Bibliographical References}\label{sec:reference}

\bibliographystyle{lrec-coling2024-natbib}
\bibliography{lrec-coling2024-example}

\begin{thebibliography}{2}
\expandafter\ifx\csname natexlab\endcsname\relax\def\natexlab#1{#1}\fi

\bibitem[{Chen et~al.(2021)Chen, wei Chang, Schlinger, Wang, and Cohen}]{chen2021ottqa}
Wenhu Chen and Ming-wei Chang and Eva Schlinger and William Wang and William Cohen. 2021.
\newblock \href {https://openreview.net/forum?id=MmCRswl1UYl} {\emph{Open Question Answering over Tables and Text}}.
\newblock Proceedings of ICLR 2021.
\newblock PID \href{https://github.com/wenhuchen/OTT-QA}{https://github.com/wenhuchen/OTT-QA}.

\bibitem[{Chen et~al.(2020)Chen, Zha, Chen, Xiong, Wang, and Wang}]{chen-etal-2020-hybridqa}
Chen, Wenhu and Zha, Hanwen and Chen, Zhiyu and Xiong, Wenhan and Wang, Hong and Wang, William Yang. 2020.
\newblock \href {https://doi.org/10.18653/v1/2020.findings-emnlp.91} {\emph{{H}ybrid{QA}: A Dataset of Multi-Hop Question Answering over Tabular and Textual Data}}.
\newblock Association for Computational Linguistics.
\newblock PID \href{https://github.com/wenhuchen/HybridQA}{https://github.com/wenhuchen/HybridQA}.

\end{thebibliography}


\begin{thebibliography}{17}
\expandafter\ifx\csname natexlab\endcsname\relax\def\natexlab#1{#1}\fi

\bibitem[{Ainslie et~al.(2020)Ainslie, Ontanon, Alberti, Cvicek, Fisher, Pham, Ravula, Sanghai, Wang, and Yang}]{ainslie-etal-2020-etc}
Joshua Ainslie, Santiago Ontanon, Chris Alberti, Vaclav Cvicek, Zachary Fisher, Philip Pham, Anirudh Ravula, Sumit Sanghai, Qifan Wang, and Li~Yang. 2020.
\newblock \href {https://doi.org/10.18653/v1/2020.emnlp-main.19} {{ETC}: Encoding long and structured inputs in transformers}.
\newblock In \emph{Proceedings of the 2020 Conference on Empirical Methods in Natural Language Processing (EMNLP)}, pages 268--284, Online. Association for Computational Linguistics.

\bibitem[{Beltagy et~al.(2020)Beltagy, Peters, and Cohan}]{Beltagy2020Longformer}
Iz~Beltagy, Matthew~E. Peters, and Arman Cohan. 2020.
\newblock Longformer: The long-document transformer.
\newblock \emph{arXiv:2004.05150}.

\bibitem[{Devlin et~al.(2019)Devlin, Chang, Lee, and Toutanova}]{devlin-etal-2019-bert}
Jacob Devlin, Ming-Wei Chang, Kenton Lee, and Kristina Toutanova. 2019.
\newblock \href {https://doi.org/10.18653/v1/N19-1423} {{BERT}: Pre-training of deep bidirectional transformers for language understanding}.
\newblock In \emph{Proceedings of the 2019 Conference of the North {A}merican Chapter of the Association for Computational Linguistics: Human Language Technologies, Volume 1 (Long and Short Papers)}, pages 4171--4186, Minneapolis, Minnesota. Association for Computational Linguistics.

\bibitem[{Huang et~al.(2022)Huang, Zhong, Liu, Gong, Jiang, and Duan}]{huang-etal-2022}
Junjie Huang, Wanjun Zhong, Qian Liu, Ming Gong, Daxin Jiang, and Nan Duan. 2022.
\newblock \href {https://aclanthology.org/2022.findings-emnlp.303} {{Mixed-modality Representation Learning and Pre-training for Joint Table-and-Text Retrieval in OpenQA}}.
\newblock In \emph{Findings of the Association for Computational Linguistics: EMNLP 2022}, pages 4117--4129, Abu Dhabi, United Arab Emirates. Association for Computational Linguistics.

\bibitem[{Karpukhin et~al.(2020)Karpukhin, Oguz, Min, Lewis, Wu, Edunov, Chen, and Yih}]{karpukhin2020dense}
Vladimir Karpukhin, Barlas Oguz, Sewon Min, Patrick Lewis, Ledell Wu, Sergey Edunov, Danqi Chen, and Wen-tau Yih. 2020.
\newblock Dense passage retrieval for open-domain question answering.
\newblock In \emph{Proceedings of the 2020 Conference on Empirical Methods in Natural Language Processing (EMNLP)}, pages 6769--6781.

\bibitem[{Kosti{\'c} et~al.(2021)Kosti{\'c}, Risch, and M{\"o}ller}]{kostic-etal-2021}
Bogdan Kosti{\'c}, Julian Risch, and Timo M{\"o}ller. 2021.
\newblock \href {https://doi.org/10.18653/v1/2021.mrqa-1.8} {{Multi-modal Retrieval of Tables and Texts Using Tri-encoder Models}}.
\newblock In \emph{Proceedings of the 3rd Workshop on Machine Reading for Question Answering}, pages 82--91, Punta Cana, Dominican Republic. Association for Computational Linguistics.

\bibitem[{Kumar et~al.(2023)Kumar, Gupta, Chemmengath, Sen, Chakrabarti, Bharadwaj, and Pan}]{kumar-etal-2023-multi}
Vishwajeet Kumar, Yash Gupta, Saneem Chemmengath, Jaydeep Sen, Soumen Chakrabarti, Samarth Bharadwaj, and Feifei Pan. 2023.
\newblock \href {https://doi.org/10.18653/v1/2023.acl-long.449} {Multi-row, multi-span distant supervision for {T}able+{T}ext question answering}.
\newblock In \emph{Proceedings of the 61st Annual Meeting of the Association for Computational Linguistics (Volume 1: Long Papers)}, pages 8080--8094, Toronto, Canada. Association for Computational Linguistics.

\bibitem[{Lei et~al.(2023)Lei, Li, Wei, He, Huang, Zhao, and Liu}]{lei-etal-2023-s3hqa}
Fangyu Lei, Xiang Li, Yifan Wei, Shizhu He, Yiming Huang, Jun Zhao, and Kang Liu. 2023.
\newblock \href {https://doi.org/10.18653/v1/2023.acl-short.147} {{S}3{HQA}: A three-stage approach for multi-hop text-table hybrid question answering}.
\newblock In \emph{Proceedings of the 61st Annual Meeting of the Association for Computational Linguistics (Volume 2: Short Papers)}, pages 1731--1740, Toronto, Canada. Association for Computational Linguistics.

\bibitem[{Li et~al.(2021)Li, Ng, Xu, Zhu, Wang, and Xiang}]{li-etal-2021-dual}
Alexander~Hanbo Li, Patrick Ng, Peng Xu, Henghui Zhu, Zhiguo Wang, and Bing Xiang. 2021.
\newblock \href {https://doi.org/10.18653/v1/2021.acl-long.315} {Dual reader-parser on hybrid textual and tabular evidence for open domain question answering}.
\newblock In \emph{Proceedings of the 59th Annual Meeting of the Association for Computational Linguistics and the 11th International Joint Conference on Natural Language Processing (Volume 1: Long Papers)}, pages 4078--4088, Online. Association for Computational Linguistics.

\bibitem[{Liu et~al.(2019)Liu, Ott, Goyal, Du, Joshi, Chen, Levy, Lewis, Zettlemoyer, and Stoyanov}]{liu2019roberta}
Yinhan Liu, Myle Ott, Naman Goyal, Jingfei Du, Mandar Joshi, Danqi Chen, Omer Levy, Mike Lewis, Luke Zettlemoyer, and Veselin Stoyanov. 2019.
\newblock Roberta: A robustly optimized bert pretraining approach.
\newblock \emph{arXiv preprint arXiv:1907.11692}.

\bibitem[{Ma et~al.(2022)Ma, Cheng, Liu, Nyberg, and Gao}]{ma-etal-2022}
Kaixin Ma, Hao Cheng, Xiaodong Liu, Eric Nyberg, and Jianfeng Gao. 2022.
\newblock \href {https://aclanthology.org/2022.findings-emnlp.392} {{Open-domain Question Answering via Chain of Reasoning over Heterogeneous Knowledge}}.
\newblock In \emph{Findings of the Association for Computational Linguistics: EMNLP 2022}, pages 5360--5374, Abu Dhabi, United Arab Emirates. Association for Computational Linguistics.

\bibitem[{Ma et~al.(2023)Ma, Cheng, Zhang, Liu, Nyberg, and Gao}]{ma-etal-2023}
Kaixin Ma, Hao Cheng, Yu~Zhang, Xiaodong Liu, Eric Nyberg, and Jianfeng Gao. 2023.
\newblock \href {https://arxiv.org/abs/2305.03130} {{Chain-of-Skills: A Configurable Model for Open-Domain Question Answering}}.
\newblock In \emph{ArXiv e-prints}.
\newblock \texttt{arXiv: 2305.03130 [cs.CL]}.

\bibitem[{Park et~al.(2023)Park, Lee, Seo, Kim, Kang, and Na}]{park2023rink}
Eunhwan Park, Sung-Min Lee, Dearyong Seo, Seonhoon Kim, Inho Kang, and Seung-Hoon Na. 2023.
\newblock Rink: reader-inherited evidence reranker for table-and-text open domain question answering.
\newblock In \emph{Proceedings of the AAAI Conference on Artificial Intelligence}, volume~37, pages 13446--13456.

\bibitem[{Qu et~al.(2021)Qu, Ding, Liu, Liu, Ren, Zhao, Dong, Wu, and Wang}]{qu-etal-2021-rocketqa}
Yingqi Qu, Yuchen Ding, Jing Liu, Kai Liu, Ruiyang Ren, Wayne~Xin Zhao, Daxiang Dong, Hua Wu, and Haifeng Wang. 2021.
\newblock \href {https://doi.org/10.18653/v1/2021.naacl-main.466} {{R}ocket{QA}: An optimized training approach to dense passage retrieval for open-domain question answering}.
\newblock In \emph{Proceedings of the 2021 Conference of the North American Chapter of the Association for Computational Linguistics: Human Language Technologies}, pages 5835--5847, Online. Association for Computational Linguistics.

\bibitem[{Robertson et~al.(2009)Robertson, Zaragoza et~al.}]{robertson2009probabilistic}
Stephen Robertson, Hugo Zaragoza, et~al. 2009.
\newblock The probabilistic relevance framework: Bm25 and beyond.
\newblock \emph{Foundations and Trends{\textregistered} in Information Retrieval}, 3(4):333--389.

\bibitem[{Wu et~al.(2020)Wu, Petroni, Josifoski, Riedel, and Zettlemoyer}]{wu-etal-2020-scalable}
Ledell Wu, Fabio Petroni, Martin Josifoski, Sebastian Riedel, and Luke Zettlemoyer. 2020.
\newblock \href {https://doi.org/10.18653/v1/2020.emnlp-main.519} {Scalable zero-shot entity linking with dense entity retrieval}.
\newblock In \emph{Proceedings of the 2020 Conference on Empirical Methods in Natural Language Processing (EMNLP)}, pages 6397--6407, Online. Association for Computational Linguistics.

\bibitem[{Zhong et~al.(2022)Zhong, Huang, Liu, Zhou, Wang, Yin, and Duan}]{zhong-etal-2022}
Wanjun Zhong, Junjie Huang, Qian Liu, Ming Zhou, Jiahai Wang, Jian Yin, and Nan Duan. 2022.
\newblock \href {https://doi.org/10.24963/ijcai.2022/629} {{Reasoning over Hybrid Chain for Table-and-Text Open Domain Question Answering}}.
\newblock In \emph{Proceedings of the Thirty-First International Joint Conference on Artificial Intelligence, IJCAI-22}, pages 4531--4537. International Joint Conferences on Artificial Intelligence Organization.

\end{thebibliography}

\section{Language Resource References}
\label{lr:ref}
\bibliographystylelanguageresource{lrec-coling2024-natbib}
\bibliographylanguageresource{languageresource}

\end{document}